\documentclass{article}

\usepackage[preprint]{corl_2026} 
\usepackage{multirow}
\usepackage{booktabs}
\usepackage{tabularx}
\usepackage{array}
\usepackage{graphicx}
\usepackage{amsmath}
\usepackage{amssymb}
\usepackage{bm}
\usepackage{algorithm}
\usepackage{algpseudocode}
\usepackage{xspace}
\usepackage{hyperref}

\newcommand{\algabbr}{TrAct\xspace}

\title{TrAct: Bridging Robot Control and Visual Prediction with Visual Tracks}

\author{
  Zhi Cao\textsuperscript{1,2,*} \quad
  Howard Ji\textsuperscript{2,*} \quad
  Kevin Zhang\textsuperscript{2,*} \quad
  Kuangzhi Ge\textsuperscript{2} \\[0.35em]
  {\bf Li Fei-Fei\textsuperscript{2} \quad
  Jiajun Wu\textsuperscript{2,\dag} \quad
  Huang Huang\textsuperscript{2,\dag}} \\[0.6em]
  {\normalfont\normalsize
  \textsuperscript{1}University of Michigan \quad
  \textsuperscript{2}Stanford University} \\[0.4em]
  {\normalfont\small \textsuperscript{*}Equal contribution. \quad
  \textsuperscript{\dag}Equal advising.}
}

\begin{document}
\maketitle

\begin{center}
    \href{https://lol-png.github.io/tract/}
    {\texttt{https://lol-png.github.io/tract/}}
\end{center}


\begin{figure*}[!htb]
    \centering
    \includegraphics[width=1\linewidth]{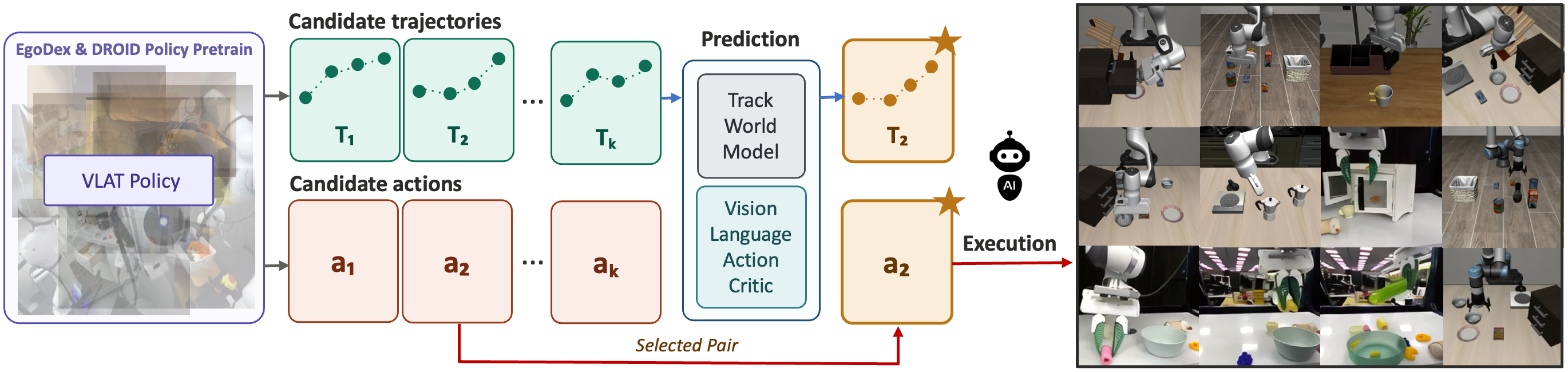}
    \caption{
\textbf{Overview of \algabbr{}.} \algabbr{} uses VLAT (left, purple), pretrained on large-scale cross-embodiment data, to propose candidate action--track pairs. TWM (middle, gray) rolls out the visual outcome for each candidate conditioned on the tracks, and VLAC (middle, green) selects the highest-reward rollout; the robot executes the paired action. Visual tracks thus serve as an intermediate interface between future prediction and robot control. We evaluate \algabbr{} on diverse simulation and physical robot tasks (right).
    }
    \label{fig:splash}
\end{figure*}

\begin{abstract}
Robot actions are inherently embodiment-specific and only weakly aligned with image-space visual changes, limiting their effectiveness as conditioning signals for robot world models. In contrast, visual tracks provide an embodiment-agnostic representation of how task-relevant points move through a scene, offering dense image-space guidance for accurate and spatially precise future video prediction. Building on this observation, we propose \textbf{\algabbr}, a world-model-based robot decision-making framework that uses visual tracks as an intermediate interface between control and prediction. \algabbr consists of three components: a Vision-Language-Action-and-Track model (VLAT) that jointly predicts candidate actions and corresponding visual tracks from the current observation and language instruction; a track-conditioned world model (TWM) that predicts future visual outcomes conditioned on the proposed tracks; and a vision-language reward model (VLAC) that scores the predicted outcomes. At inference time, VLAT generates candidate action--track pairs, TWM rolls out their visual consequences conditioned on the tracks, and VLAC selects the track whose predicted outcome best satisfies the instruction; the action paired with the selected track is then executed by the robot. Experiments on the proposed LIBERO-INTEGRAL benchmark and real-world Franka manipulation show that \algabbr improves success from 27\% to 55\% in simulation and 49\% to 76\% on real-world tasks compared to the strong VLA baseline $\pi_{0.5}$. Furthermore, TWM consistently improves video prediction quality over AWM. These results suggest that visual tracks provide an effective shared interface between robot control and visual prediction, enabling more accurate world modeling and stronger robot generalization.
\end{abstract}


\section{Introduction}
\label{sec:intro}

Robot world models predict scene evolution under candidate decisions~\citep{ebert2018visualforesightmodelbaseddeep, du2023learning, yang2024learninginteractiverealworldsimulators}, while policies produce executable actions. Recent work combines them by using a world model to evaluate candidate actions and select the predicted outcome that best satisfies the task objective~\citep{qi2026inference}. However, robot actions, which are typically used to connect the policy and the world model, are a weak interface between control and prediction: actions are embodiment-specific, and their visual effects depend strongly on scene geometry, object configuration, and contact dynamics. Thus, action-conditioned world models must infer dense visual futures from sparse, robot-specific commands.
Visual tracks provide a natural alternative. They describe how task-relevant points move through image space, capturing object motion, contact regions, and interaction geometry, encouraging generalization. Because different robots can induce similar scene-point motion with different commands, tracks offer a compact, largely embodiment-agnostic representation for both policy learning and visual prediction.

We propose \textbf{TrAct} (\textbf{Tr}ack \textbf{\&} \textbf{Act}), a framework that uses visual tracks as a shared interface between policy learning and world-model-based decision making. \algabbr consists of three components: (1) a \emph{Vision-Language-Action-and-Track} model (VLAT) that predicts candidate actions and corresponding visual tracks from the current observation and language instruction; (2) a \emph{track-conditioned world model} (TWM) that predicts future visual outcomes from proposed tracks; and (3) a \emph{vision-language reward model} (VLAC) that scores predicted outcomes against the task instruction. During inference, VLAT proposes multiple paired action--track candidates, TWM predicts the visual outcome induced by each track, and VLAC evaluates these outcomes with respect to the task goal. \algabbr{} then selects the highest-scoring predicted track outcome and executes its paired robot action.

We evaluate \algabbr{} on LIBERO-INTEGRAL, which combines task variations from LIBERO-PRO and LIBERO-Plus and includes UR5 cross-embodiment transfer tasks, as well as on a physical Franka Panda robot. \algabbr{}'s track-conditioned world model (TWM) consistently outperforms action-conditioned world model baselines (AWM) for future prediction in both settings. This improved prediction translates into more effective world-model-based action selection, improving success from 49\% to 55\% on LIBERO-INTEGRAL and from 66\% to 76\% on real-world tasks compared to AWM-based selection. Compared with the strong VLA baseline $\pi_{0.5}$~\cite{intelligence2025pi05visionlanguageactionmodelopenworld}, \algabbr{} improves success from 27\% to 55\% on LIBERO-INTEGRAL and from 49\% to 76\% on real-world tasks. These results show that visual tracks provide an effective intermediate representation between planning and execution, improving both future prediction and task completion.


Our contributions are:
\begin{itemize}
    \item We propose 2D point tracks as an intermediate representation between robot policies and world models, improving both planning and future prediction.
    \item We introduce \algabbr{}, a framework that uses a VLAT policy to jointly predict visual tracks and actions, and conditions a world model on tracks for reward-guided action selection.
    \item We validate \algabbr{} in simulation and on physical robots, showing that TWM improves future prediction over action-conditioned alternatives and that TWM-based action selection increases downstream task success.
\end{itemize}

\section{Related Work}
\label{sec:rw}

\textbf{Embodied Video Generation World Models.}
Video-generation world models learn neural simulators of physical interaction from visual data. Internet-scale models such as Sora~\citep{openai2024sora,openai2025sora2} and Genie~\citep{bruce2024geniegenerativeinteractiveenvironments} exhibit emergent physical understanding, but lack the controllability needed for precise robot control. Embodied world models improve robotic motion consistency by conditioning video prediction on language, actions, proprioception, multimodal inputs, hand-centric priors, or visual action prompts~\citep{yang2024learninginteractiverealworldsimulators,xiang2024pandorageneralworldmodel,smith2020avidlearningmultistagetasks,chi2025wowworldomniscientworld,li2025multimodalactionconditionedvideo,li2025handihandcentrictextandimageconditioned,wang2025preciseactiontovideogenerationvisual}. Yet action conditioning remains poorly aligned with image-space change: actions are low-dimensional and embodiment-specific, while their visual effects depend on geometry, object state, and contact. This underdetermined action-to-pixel mapping causes poor action following; \citet{liu2026worldvlaloopclosedlooplearningvideo} find that such models often ignore erroneous actions and hallucinate success from visual priors. In contrast, \algabbr{} conditions on visual tracks, which directly specify point motion and provide dense spatial guidance for video prediction.

\textbf{Visual Tracks and Embodiment-Agnostic Action Representations.}
Structured motion representations such as 2D point tracks and optical flow separate motion from visual rendering, reducing the complexity of both prediction and control. Enabled by long-horizon trackers~\citep{karaev2024cotrackerbettertrack,doersch2023tapirtrackingpointperframe}, one line of work treats such motion as an intermediate \emph{policy} representation, decoding predicted tracks or flow into executable actions and thereby transferring across embodiments and from human video to robots~\citep{wen2023anypoint,xu2024flow,bharadhwaj2024track2act,yuan2025generalflow}; related work instead consumes predicted visual representations for policy learning or supplies trajectory-like prompts to video generators~\citep{hu2025videopredictionpolicygeneralist,chi2025mindlearningdualsystemworld,geng2025motionpromptingcontrollingvideo,zheng2025tracevlavisualtraceprompting}. A parallel line replaces embodiment-specific commands with embodiment-agnostic motion as the \emph{conditioning signal of a world model}, on the premise that low-dimensional robot commands underdetermine visual change~\citep{wang2025latentpolicysteering,huang2026pointworld}. Together, these establish visual motion as an effective interface on both sides of the control--prediction loop. Across both lines, however, motion is used to produce or condition a prediction, while the predicted future itself is evaluated in latent or geometric space, if at all. \algabbr{} instead uses tracks as the interface that renders a predicted future assessable against the task instruction.

\textbf{Vision-Language-Action Models and Closed-Loop Control.}
Vision-Language-Action (VLA) models combine semantic reasoning with visual-motor control and have become a central paradigm for generalist robot policies. Representative models span discretized action tokens and continuous action decoders~\citep{brohan2023rt2visionlanguageactionmodelstransfer,kim2024openvlaopensourcevisionlanguageactionmodel,octo2024}, with $\pi_0$~\citep{black2026pi0visionlanguageactionflowmodel} introducing flow matching for continuous action generation. Despite strong priors, open-loop VLAs remain brittle under distribution shift, motivating closed-loop extensions such as VLMPC~\citep{zhao2024vlmpcvisionlanguagemodelpredictive}, VLAC~\citep{zhai2025visionlanguageactioncriticmodelroboticrealworld}, and RT-Trajectory~\citep{gu2023rttrajectoryrobotictaskgeneralization}. \algabbr{} extends this direction by augmenting a flow-matching VLA with track prediction and using a track-conditioned world model for reward-guided action selection, leveraging video-model priors while grounding selection in predicted image-space dynamics.

\begin{figure*}[t]
    \centering
    \includegraphics[width=1\linewidth]{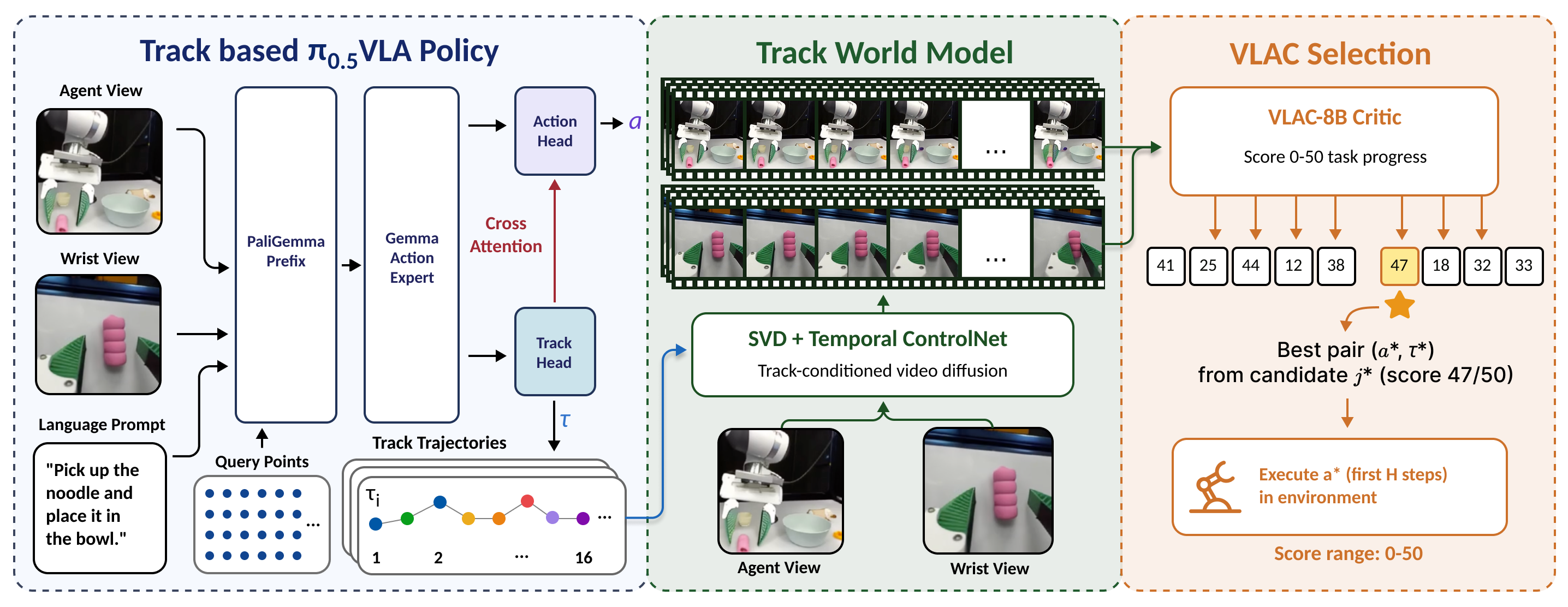}
    \caption{\textbf{Overview of the TrAct framework.} Given an observation $o_t$ and language instruction $l$, the VLAT model jointly predicts $K$ candidate actions and corresponding visual tracks. Each set of predicted tracks conditions a world model to generate future video frames. A VLM-based reward model scores the predicted outcomes and selects the action whose rollout best fits the instruction.}
    \label{fig:pipeline}
    \vspace{-1.5em}
\end{figure*}

\section{Methods}
\label{sec:methods}

Given observation $o_t$ and language instruction $l$, our goal is to select action $a_t$ that maximizes task success rate. As illustrated in Figure~\ref{fig:pipeline}, \algabbr decomposes this into three stages: (1)~generating candidate action--track pairs, (2)~rolling out their visual consequences via a track-conditioned world model, and (3)~scoring the predicted outcome proposals with a vision-language reward model. We employ InternVL2~\cite{chen2024internvlscalingvisionfoundation} as a reward model following the VLAC framework~\cite{zhai2025visionlanguageactioncriticmodelroboticrealworld}.

\subsection{Policy Architecture}
\textbf{Vision-Language-Action-and-Track Policy.}
\label{sec:tract}
We build our Vision Language Action and Track model (VLAT) on top of the pre-trained flow-matching VLA policy $\pi_{0.5}$~\cite{intelligence2025pi05visionlanguageactionmodelopenworld}. We modify its action head to jointly output executable robot actions and 2D point tracks:
\begin{equation}
    \pi_\theta(o_t, l) \;\longrightarrow\; \bigl\{(a_i,\;\tau_i)\bigr\}_{i=1}^{K},
\end{equation}
where $a_i \in \mathbb{R}^{H \times d_a}$ is a predicted action chunk of horizon $H$ and action dimension $d_a$, and $\tau_i \in \mathbb{R}^{N \times H \times 2}$ is a set of $N$ predicted 2D point tracks over the same horizon. Following $\pi_{0.5}$, the PaliGemma backbone encodes the observations and instruction as a prefix, and the Gemma action expert denoises the output tokens via flow matching. We extend the action expert with a track stream: track and action tokens use separate projections but share the action-expert trunk. Track tokens precede action tokens, so action tokens attend to the track tokens but not vice versa. Tracks and actions are thus denoised jointly, yielding coupled action--track candidates.

The VLAT policy is optimized using a joint flow-matching objective:
\begin{equation}
    \mathcal{L}_{\text{VLAT}}
    =
    \mathcal{L}_{\text{flow}}(a,a^*)
    +
    \mathcal{L}_{\text{flow}}(\tau,\tau^*),
\end{equation}
where $a^*$ and $\tau^*$ denote the ground-truth actions and tracks.
We predict tracks for two types of points: (1)~seven key points on the robotic gripper, and (2)~a uniform $5{\times}5$ grid of scene points. We use these asymmetrically across camera views. In the agent view, VLAT predicts only the gripper points. Predicting environment dynamics is challenging, so rather than having the policy explicitly specify future scene motion, we offload this burden to the world model, which infers how the scene responds to the predicted end-effector motion. In the wrist view, the camera moves with the gripper, so VLAT additionally predicts the scene-grid points to capture camera motion relative to the environment. This yields 7 agent-view and 32 wrist-view tracks per timestep.

\textbf{Track-Conditioned World Model.}
We adopt Stable Video Diffusion (SVD)~\cite{blattmann2023stablevideodiffusionscaling} as the video generation backbone. To condition generation on the predicted tracks, we augment SVD with a ControlNet branch~\cite{zhang2023addingconditionalcontroltexttoimage} that encodes track coordinates as spatial control maps. Specifically, the tracks are rasterized into blurred heatmaps, encoded by the SVD VAE, and injected into the UNet through a temporal ControlNet. Given observation $o_t$ and predicted tracks $\tau_i$, the world model generates a future video:
\begin{equation}
    \hat{v}_i = p_\phi(o_t, \tau_i) \in \mathbb{R}^{H \times C \times W_{\text{img}} \times H_{\text{img}}},
\end{equation}
where $\hat{v}_i$ is the predicted video rollout for candidate $i$.
To support multi-view observations, we render view-specific tracks into separate ControlNet channels: agent-view tracks in red and wrist-view tracks in blue. This lets the world model distinguish view-specific motion while maintaining cross-view temporal consistency. This provides denser spatial guidance than raw action vectors.

\textbf{Baseline: Action-Conditioned World Model.}
For baseline comparison, we also train an action-conditioned variant that replaces track conditioning with action conditioning. Given a predicted action chunk $a_i$, we first encode it using an MLP into the CLIP-L/14 feature space~\cite{radford2021learningtransferablevisualmodels}. To support multi-view observations, we introduce a learnable camera-view embedding indicating agent-view or wrist-view actions. The view embedding is concatenated to the encoded action latent. The resulting view-aware action representation is then injected into SVD's UNet layers via cross-attention. This design provides the action-conditioned world model baseline with the same multi-view information. Both world models are trained on ground-truth conditioning: TWM on ground-truth tracks (projected end-effector gripper tracks and CoTracker grid tracks) and AWM on ground-truth actions. At inference, both are conditioned on VLAT's predictions.

\textbf{VLAC Reward Model.}
To score predicted video results, we use the Vision-Language-Action-Critic (VLAC) model~\citep{zhai2025visionlanguageactioncriticmodelroboticrealworld}. Given a predicted video and the instruction, VLAC judges task progress on a reward scale from 0 to 50. We adopt VLAC directly from the original authors, only finetuning it on our task data, and refer readers to~\citep{zhai2025visionlanguageactioncriticmodelroboticrealworld} for implementation details.

\subsection{Pretraining}
\label{sec:training}
\paragraph{VLAT.}
VLAT is pretrained on large-scale real-world robot datasets, including 76K DROID~\citep{khazatsky2025droidlargescaleinthewildrobot} trajectories and 150K EgoDex~\citep{hoque2026egodexlearningdexterousmanipulation} trajectories (sampled from approximately half of the full EgoDex dataset) using a 1:2 dataset mixing ratio. For track supervision, we use seven predefined gripper points corresponding to fixed 3D offsets on the end-effector mesh. These points are projected into each frame using robot poses and camera calibration, providing ground-truth gripper tracks across all views. To capture scene dynamics, we additionally sample a uniform $5\times5$ grid of background points and track them over time using CoTracker~\citep{karaev2024cotrackerbettertrack}. Pretraining on diverse real-world manipulation data encourages the model to learn embodiment-agnostic visual motion representations that transfer across robot embodiments, camera viewpoints, and environments. All results in this paper use the DROID+EgoDex pretrained model trained for 30K steps with a batch size of 64 on 4 H100 GPUs.

\paragraph{World Model.}
Both the track-conditioned (TWM) and action-conditioned (AWM) world models are pretrained on 76K DROID trajectories for 30K steps with a batch size of 64. During finetuning, both variants are trained on the same LIBERO and real-world datasets to ensure a fair comparison. Additional dataset statistics, training details, and ablations, including experiments using a larger DROID+BridgeData V2~\citep{walke2024bridgedatav2datasetrobot}+EgoDex pretraining mixture, are provided in the appendix.

\subsection{Inference}
\label{sec:inference}
At inference, \algabbr{} runs the three stages of Figure~\ref{fig:pipeline}. First, VLAT proposes $K$ candidate action--track pairs $\{(a_i,\tau_i)\}_{i=1}^{K}$. To make these candidates diverse, we apply Temperature-Scaled Resampling (TSR) during flow-matching sampling: we scale the initial noise magnitude to control diversity and resample from the learned policy distribution. Second, the track-conditioned world model rolls out a predicted video $\hat{v}_i$ for each of the $K$ candidates, conditioned on its track $\tau_i$. Third, VLAC scores the $K$ rollouts and selects the best:
\begin{equation}
    i^* = \arg\max_{i \in \{1,\ldots,K\}} \; R_\psi(\hat{v}_i, \, l),
\end{equation}
where $R_\psi$ denotes the reward model. The action of the selected pair is executed by the robot. We use $K{=}20$ candidates in simulation and $K{=}16$ in the real world, with a noise scale of 2 and TSR $k{=}3$ (simulation) and $k{=}4$ (real world). Per-component latency is reported in Table~\ref{tab:latency} (Appendix).

\section{Simulation Experiments}
\label{sec:exp}

We compare four policy variants. \textbf{$\pi_{0.5}$} is the vanilla $\pi$0.5 baseline policy. \textbf{VLAT} jointly predicts visual tracks and executable actions without action selection. \textbf{\algabbr} further incorporates track-conditioned world-model action selection, while \textbf{VLAT+AWM} replaces track conditioning with action conditioning. Each comparison isolates one component: $\pi_{0.5}$ vs.\ VLAT measures the effect of joint action--track prediction; VLAT vs.\ VLAT+AWM measures world-model-based action selection; and VLAT+AWM vs.\ \algabbr{} changes only the world-model conditioning, holding the candidate generator, VLAC critic, and selection rule fixed. 

\subsection{LIBERO-INTEGRAL Benchmark}
Since standard LIBERO benchmarks are nearly saturated, we introduce \textbf{LIBERO-INTEGRAL} to evaluate policy generalization under distribution and embodiment shifts. LIBERO-INTEGRAL includes 20 tasks: 10 robustness tasks and 10 cross-embodiment tasks.
The robustness suite combines object, position, and task generalization from LIBERO-PRO~\cite{zhou2025libero} with the Different 3D Camera Viewpoints and Robot Initialization settings from LIBERO-Plus~\cite{fei2025libero}. We select two representative tasks per variation category and use the hardest available difficulty level for each LIBERO-Plus task.
For cross-embodiment evaluation, we replace the default Franka Panda with a UR5 while keeping task specifications and scene configurations fixed. We construct UR5 variants of the 10 LIBERO-10 tasks and evaluate performance under this embodiment shift.

\begin{table*}[!t]
\centering
\small
\setlength{\tabcolsep}{3.5pt}
\renewcommand{\arraystretch}{1.12}
\caption{
Video generation quality of action-conditioned and track-conditioned world models. Results are reported for real-world and simulation scenes under agent-view and wrist-view cameras. Act. denotes action-conditioned world model, and Track denotes track-conditioned world model. $\uparrow$ indicates higher is better and $\downarrow$ indicates lower is better. Best results for each scene are in bold.}
\label{tab:video_gen_quality}
\vspace{4pt}  
\begin{tabular*}{\textwidth}{@{\extracolsep{\fill}}lcc cc cc cc@{}}
\toprule
\multirow{2}{*}{\textbf{Metric}}
& \multicolumn{2}{c}{\textbf{Real / Agent}}
& \multicolumn{2}{c}{\textbf{Real / Wrist}}
& \multicolumn{2}{c}{\textbf{Sim / Agent}}
& \multicolumn{2}{c}{\textbf{Sim / Wrist}} \\
\cmidrule(lr){2-3}
\cmidrule(lr){4-5}
\cmidrule(lr){6-7}
\cmidrule(lr){8-9}
& \textbf{Act.} & \textbf{Track}
& \textbf{Act.} & \textbf{Track}
& \textbf{Act.} & \textbf{Track}
& \textbf{Act.} & \textbf{Track} \\
\midrule
PSNR $\uparrow$
& 20.70 & \textbf{21.27}
& 16.89 & \textbf{20.62}
& 15.12 & \textbf{24.51}
& 16.25 & \textbf{22.71} \\

SSIM $\uparrow$
& 0.767 & \textbf{0.802}
& 0.621 & \textbf{0.738}
& 0.482 & \textbf{0.843}
& 0.668 & \textbf{0.822} \\

LPIPS $\downarrow$
& 0.126 & \textbf{0.111}
& 0.372 & \textbf{0.225}
& 0.438 & \textbf{0.106}
& 0.452 & \textbf{0.182} \\

FID $\downarrow$
& 105 & \textbf{95}
& 176 & \textbf{130}
& 248 & \textbf{156}
& 222 & \textbf{92} \\

FVD $\downarrow$
& 148 & \textbf{105}
& 246 & \textbf{140}
& 129 & \textbf{38}
& 207 & \textbf{162} \\
\bottomrule
\end{tabular*}
\vspace{-1em}
\end{table*}

\paragraph{Model Finetuning.}
For track supervision, we use seven predefined gripper points defined by fixed 3D offsets on the end-effector mesh. Robot poses and camera calibration project these points into each frame to obtain ground-truth gripper tracks. To capture scene dynamics, we also sample a uniform $5\times5$ grid of wrist-view background points and track them with CoTracker~\cite{karaev2024cotrackerbettertrack}. For $\pi_{0.5}$, we use the official LIBERO checkpoint provided by the $\pi_{0.5}$ authors. All other models are finetuned on 2K LIBERO episodes. VLAT is trained for 5K steps with a batch size of 32 to jointly predict 16-step action and track chunks. The track and action-conditioned world models are each finetuned for 25K steps with batch size 16. Following VLAC~\cite{zhai2025visionlanguageactioncriticmodelroboticrealworld}, we finetune the reward model on the same data using predicted rollouts and task completion rewards for 5K steps with a batch size of 16.

\subsection{Video Generation Results}

\begin{figure}[!htb]
    \centering
    \includegraphics[width=\columnwidth]{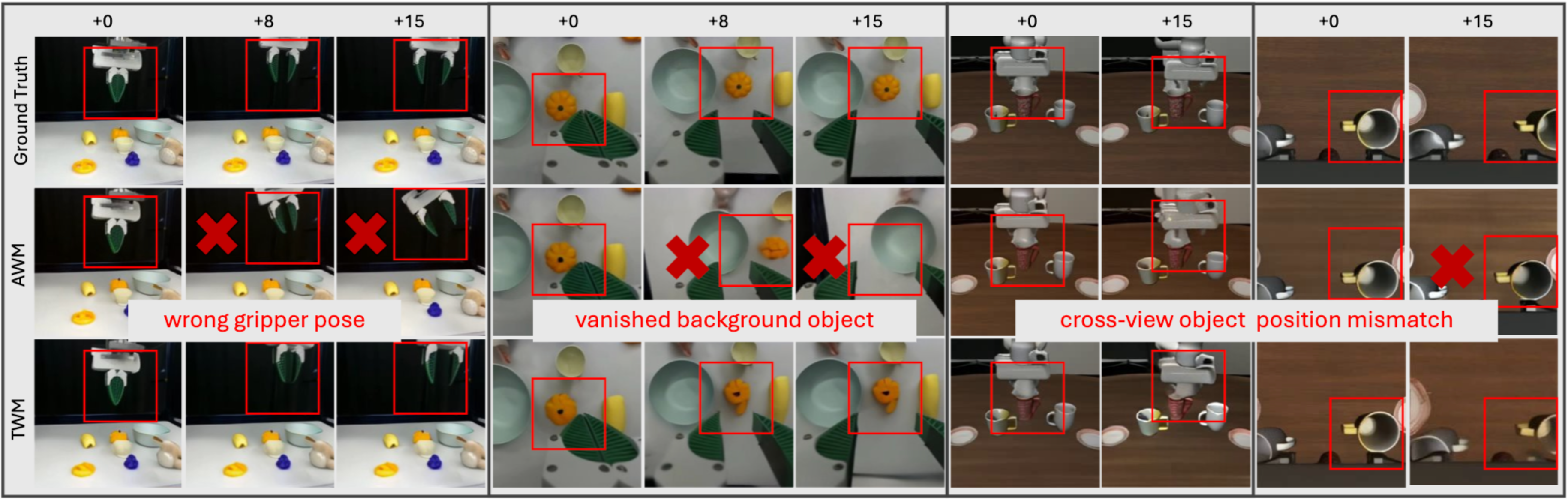}
    \caption{\textbf{Qualitative video comparison.} TWM produces more spatially accurate and motion-coherent frames than AWM across both agent-view and wrist-view cameras.}
    \label{fig:video_quality_comp}
    \vspace{-1.5em}
\end{figure}

To assess the future prediction quality, we compare the track-conditioned world model (TWM) against the action-conditioned baseline (AWM). As shown in Figure 3, TWM produces more spatially accurate and temporally coherent predictions across both agent-view and wrist-view observations, while AWM often exhibits incorrect gripper poses, disappearing objects, and cross-view object position inconsistencies. We evaluate generated videos from both cameras using standard video quality metrics. We use 100 generated videos from evaluation tasks on LIBERO-INTEGRAL. Results are shown in Table~\ref{tab:video_gen_quality}. TWM outperforms AWM on all five metrics across both camera views. The gap is large in simulation, where TWM achieves a PSNR of 24.51 vs. 15.12 on agent view and reduces LPIPS from 0.438 to 0.106, suggesting that track conditioning provides substantially richer spatial guidance than action conditioning for video generation across domains.

\subsection{Robot Experiment Results}

We first evaluate \algabbr{} on the four standard LIBERO benchmark suites and then on \textbf{LIBERO-INTEGRAL} to evaluate robustness under distribution shifts and cross-embodiment transfer. Standard LIBERO consists of four suites with 10 tasks each. For all simulation experiments, each task is evaluated using 10 rollout episodes, and success rates are averaged across rollouts. All policies predict and execute action chunks with a horizon of 16 steps during evaluation.

\begin{table}[!t]
{\centering
\small
\setlength{\tabcolsep}{4.5pt}
\caption{Success rates (\%) on four standard LIBERO benchmark suites. Each suite contains 100 evaluation episodes. The best result in each column is shown in bold.}
\label{tab:libero_standard_results}
\vspace{4pt}
\begin{tabular*}{\columnwidth}{@{\extracolsep{\fill}}lccccc@{}}
\toprule
\textbf{Method}
& \textbf{LIBERO-10}
& \textbf{Goal}
& \textbf{Object}
& \textbf{Spatial}
& \textbf{Avg.} \\
\midrule
$\pi_{0.5}$  & \textbf{94.0} & 96.0  & 98.0  & 99.0  & 96.8 \\
VLAT         & \textbf{94.0} & 98.0  & \textbf{100.0} & \textbf{100.0} & 98.0 \\
VLAT+AWM     & \textbf{94.0} & 98.0  & \textbf{100.0} & \textbf{100.0} & 98.0 \\
\algabbr{}   & \textbf{94.0} & \textbf{100.0} & 99.0 & \textbf{100.0} & \textbf{98.3} \\
\bottomrule
\end{tabular*}\par}

\vspace{10pt}

{\centering
\small
\setlength{\tabcolsep}{6.5pt}
\renewcommand{\arraystretch}{1.10}
\caption{
Results on LIBERO-INTEGRAL. Scores denote average task success rates. LIBERO-INTEGRAL combines robustness variations from LIBERO-PRO and LIBERO-Plus, and cross-embodiment transfer with a UR5 robot. Best results in each column are bolded.
}
\label{tab:libero_integral_results}
\vspace{4pt}
\begin{tabular*}{\textwidth}{@{\extracolsep{\fill}}lccccccc@{}}
\toprule
\textbf{Method}
& \textbf{Swap}
& \textbf{Object}
& \textbf{Task}
& \textbf{Camera}
& \textbf{RobotInit}
& \textbf{Cross-Emb.}
& \textbf{Avg.} \\
\midrule
$\pi_{0.5}$
& 0.40 & 0.35 & 0.15 & 0.45 & 0.45 & 0.17 & 0.27 \\

VLAT
& 0.50 & 0.40 & 0.35 & 0.45 & 0.55 & 0.42 & 0.44 \\

VLAT+AWM
& \textbf{0.65} & 0.50 & \textbf{0.50} & 0.45 & 0.55 & 0.44 & 0.49 \\

\algabbr{}
& \textbf{0.65} & \textbf{0.60} & \textbf{0.50} & \textbf{0.60} & \textbf{0.65} & \textbf{0.50} & \textbf{0.55} \\
\bottomrule
\end{tabular*}\par}
\end{table}

\paragraph{Standard LIBERO Suites.}
Table~\ref{tab:libero_standard_results} reports results on the four standard LIBERO suites, where all methods achieve strong performance. Compared with $\pi_{0.5}$ (96.8\% average success rate), VLAT improves the average success rate to 98.0\%, demonstrating that jointly predicting visual tracks and actions preserves policy performance on the original benchmark. Incorporating world-model-based action selection further improves performance, with \algabbr{} achieving the best overall result of 98.3\% and VLAT+AWM achieving 98.0\%. The relatively small gains are expected given the near-saturated performance of existing methods on these tasks, which motivates the harder LIBERO-INTEGRAL evaluation below.

\paragraph{LIBERO-INTEGRAL.}
Table~\ref{tab:libero_integral_results} reports results on the proposed LIBERO-INTEGRAL benchmark, while the complete per-task list is provided in Table~\ref{tab:libero_integral_results_app} in the Appendix. Compared with $\pi_{0.5}$, all proposed variants substantially improve generalization. VLAT raises the overall average success rate from 0.27 to 0.44, suggesting that jointly predicting visual tracks and robot actions improves robustness under distribution shifts. Adding world-model-based action selection further improves performance: VLAT+AWM reaches 0.49, while \algabbr{} achieves the best overall result of 0.55. Across three evaluation seeds, \algabbr{} reaches $0.547\pm0.003$ versus $0.490\pm0.006$ for VLAT+AWM, with non-overlapping 95\% confidence intervals (Appendix~\ref{sec:seeds_and_k}).

\algabbr{} performs best across individual variations, achieving the highest success rates on Object (0.60), Camera (0.60), RobotInit (0.65), and Cross-Embodiment (0.50) benchmarks. Compared with VLAT, \algabbr{} improves performance across all categories, and VLAT+AWM matches or improves on every category, showing the value of predictive evaluation during decision making. Replacing action conditioning with track conditioning yields consistent gains, especially under Object (0.60 vs.\ 0.50), Camera (0.60 vs.\ 0.45), RobotInit (0.65 vs.\ 0.55), and Cross-Embodiment (0.50 vs.\ 0.44) shifts.
These results support the central hypothesis that visual tracks provide a robust interface between robot control and visual prediction, improving generalization under observation, environment, and embodiment changes.

\vfill











\begin{table*}[t]
\centering
\small
\caption{Success rates on five unseen physical robot manipulation tasks, each evaluated under a regular training background (\texttt{R}) and a visually different non-training background (\texttt{D}). We compare five models. The best result for each row is shown in bold.}
\label{tab:physical_results}
\vspace{4pt}
\begin{tabular}{lcccccc}
\toprule
\textbf{Task}
& \textbf{Bg.}
& \textbf{$\pi_{0.5}$}
& \textbf{$\pi_{0.5}$+VLAC}
& \textbf{VLAT}
& \textbf{VLAT+AWM}
& \textbf{\algabbr{}} \\
\midrule
Put the green block in the bowl
& \texttt{R} & 0.5 & 0.7 & 0.6 & \textbf{0.7} & \textbf{0.7} \\
& \texttt{D} & 0.4 & 0.5 & 0.4 & 0.5 & \textbf{0.7} \\
\midrule
Put the half pink noodle in the bowl
& \texttt{R} & 0.6 & 0.9 & 0.8 & \textbf{1.0} & \textbf{1.0} \\
& \texttt{D} & 0.6 & 0.8 & 0.6 & \textbf{0.8} & \textbf{0.8} \\
\midrule
Put the hotdog bun in the bowl
& \texttt{R} & 0.6 & 0.8 & 0.6 & \textbf{0.8} & \textbf{0.8} \\
& \texttt{D} & 0.4 & 0.6 & 0.5 & \textbf{0.6} & \textbf{0.6} \\
\midrule
Close the cabinet door
& \texttt{R} & 0.5 & 0.6 & 0.5 & 0.6 & \textbf{0.8} \\
& \texttt{D} & 0.3 & 0.4 & 0.4 & 0.4 & \textbf{0.7} \\
\midrule
Pour the pasta into the pan
& \texttt{R} & 0.5 & 0.7 & 0.6 & 0.7 & \textbf{0.8} \\
& \texttt{D} & 0.5 & 0.5 & 0.5 & 0.5 & \textbf{0.7} \\
\midrule
\textbf{Average}
& & 0.49 & 0.65 & 0.55 & 0.66 & \textbf{0.76} \\
\bottomrule
\end{tabular}
\end{table*}

\section{Physical Robot Experiments}
\label{sec:physical_exp}
For physical experiments, we compare the same model variants as in the simulation experiments, and \textbf{$\pi_{0.5}$ + VLAC}, which samples candidate actions from $\pi_{0.5}$, rolls them out with the action-conditioned world model (AWM), and selects with VLAC. This additional baseline isolates the candidate generator: $\pi_{0.5}$+VLAC vs.\ VLAT+AWM changes only the policy that proposes candidates, holding the world model, VLAC critic, and selection rule fixed.

\begin{figure}[!b]
    \centering
    \includegraphics[width=\columnwidth]{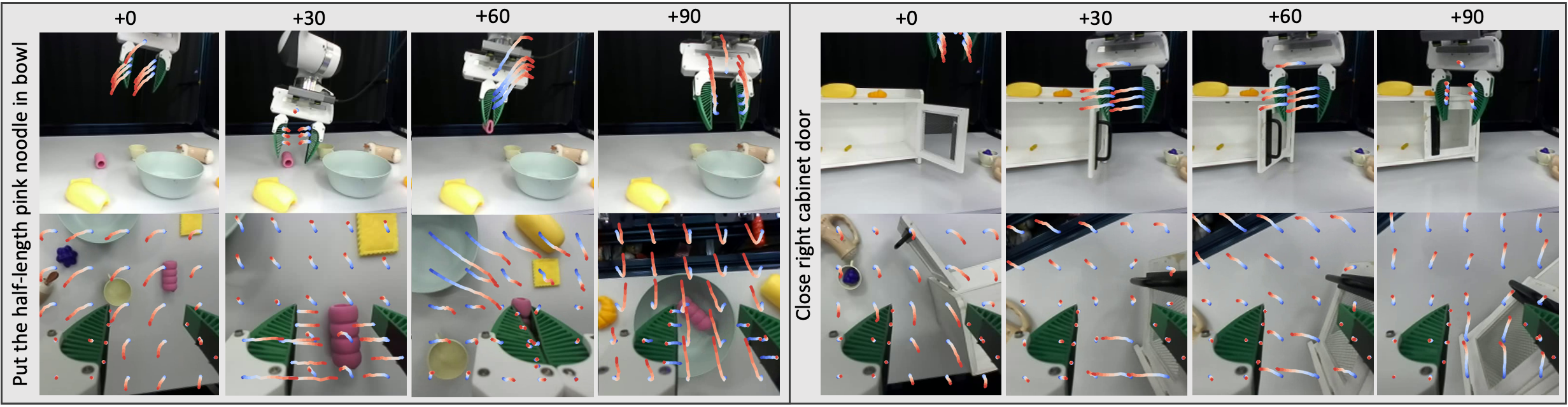}
    \caption{\textbf{Rollout Demonstrations of two tasks:} put the half-length pink noodle in the bowl (left) and close the right cabinet door (right). Each step is overlaid with track predictions.}
    \label{fig:rollout_tracks}
    \vspace{-2em}
\end{figure}

\subsection{Physical Setup}
We used a Franka Emika Panda equipped with two RGB cameras (one agent-view and one wrist-view camera). The workspace is a blank tabletop with objects placed as needed. We collected 400 demonstrations at 15 Hz for four tasks: put the pink noodle in the bowl, put the bread in the bowl, close the fully open right cabinet door, and pour the pasta into the pan with a fixed background. 

We evaluated five manipulation tasks that are unseen during training: put the half-length pink noodle in the bowl, put the hotdog bun in the bowl, put the green brick in the bowl, close the partially open right cabinet door, and pour the angled scooper's pasta into the pan. Each task is tested under two background conditions: the original training background (\texttt{reg.\ back}) and a visually distinct background (\texttt{diff back}), allowing us to measure robustness to visual domain shift on top of task-level generalization. The full setup and task selection are detailed in the Appendix.

\subsection{Model Finetuning}
The training data is preprocessed following the same method as sim. All policies are finetuned on the 400 demonstrations we collected. $\pi_{0.5}$ and VLAT are finetuned for 5k steps at batch size 32. For both action-conditioned and track-conditioned world models, we finetune for 25k steps with a batch size of 16. The base VLAC model from ~\citep{zhai2025visionlanguageactioncriticmodelroboticrealworld} is finetuned for 4k steps at batch size 16. 


\subsection{Video Generation Results}
Similar to the simulation environment, we assess the future prediction quality between the track-conditioned world model (TWM) and the action-conditioned world model (AWM).
We evaluate generated videos from both cameras using standard video quality metrics. We use 100 generated videos from evaluation rollout. Results are shown in Table~\ref{tab:video_gen_quality}. Consistent with the simulation experiments, TWM outperforms AWM on all five metrics across both camera views. On real-world tasks, the improvement is especially large on wrist view, where LPIPS improves from 0.372 to 0.225 and FVD from 246 to 140. Figure~\ref{fig:video_quality_comp} shows this qualitatively, where action-conditioned predictions are visibly less spatially accurate than track-conditioned ones.

\subsection{Robot Experiment Results}

During physical evaluation, each policy call generates a 16-step action chunk, and we execute all 16 steps. In baselines with action selection, we use 16 candidates, generating diverse samples using the same strategy as sim. Each task is evaluated over 10 episodes. Table~\ref{tab:physical_results} reports success rates on all five OOD unseen tasks under both background conditions.

Example rollouts and VLAT track predictions are shown in Figure~\ref{fig:rollout_tracks}. Both world-model variants improve over VLAT alone (0.55), confirming the value of VLM-based action selection. Track conditioning is consistently stronger: TrAct averages 0.76 versus 0.66 for VLAT+AWM. The gap widens under visual domain shift: on the training background, \algabbr{} averages 0.82 versus 0.76 for VLAT+AWM, whereas on the unseen background, \algabbr{} averages 0.70 versus 0.56. These results suggest that action-conditioned video generation degrades under visual shift, whereas track conditioning anchors the world model to image-space motion that remains robust to background changes, consistent with Table~\ref{tab:video_gen_quality} and Figure~\ref{fig:video_quality_comp}.

\section{Conclusion}
\label{sec:conclusion}

We presented \algabbr, a framework that uses visual tracks as a shared interface between robot policy learning and world-model-based decision making. By jointly predicting actions and visual tracks, and conditioning a world model on the predicted tracks rather than on raw actions, \algabbr enables reward-guided action selection grounded in image-space dynamics. Experiments on the LIBERO benchmark and on a physical Franka Panda robot show that track-conditioned world modeling consistently outperforms action-conditioned alternatives, with the largest gains under visual domain shift. These results demonstrate that visual tracks provide a practical and effective bridge between embodiment-specific control and embodiment-agnostic visual prediction.


\textbf{Limitations.} Inference is the main practical cost. Each planning step produces a 16-step action chunk rather than a single low-level control command, and takes approximately 2.4\,s on L40S GPUs; video-model denoising dominates (1810\,ms), while candidate generation and reward scoring together take under 0.7\,s (Table~\ref{tab:latency}, Appendix). As in any propose--predict--select pipeline, compute grows with $K$; we mitigate latency by rolling out candidates in parallel, and the diffusion sampler, not the track interface, is the bottleneck, addressable through few-step samplers or distillation. The uniform $5{\times}5$ grid could be replaced by adaptive sampling of task-relevant regions. Extending to broader manipulation domains and longer-horizon tasks, and developing standardized benchmarks for comparing track-based approaches (e.g., ATM), are natural next steps.

\clearpage


\bibliography{example}  

\clearpage
\appendix
\section{Pseudo-code for \algabbr{} Inference Procedure}
\label{sec:pseudo_code}

Algorithm~\ref{alg:tract_inference} summarizes the inference procedure of \algabbr{}. At each decision step, VLAT proposes K diverse action--track candidates, the track-conditioned world model imagines their visual consequences, and VLAC selects the action chunk whose predicted rollout best satisfies the instruction.

\begin{algorithm}[h!]
\caption{\algabbr{} Inference for WM-Guided Action Selection}
\label{alg:tract_inference}
\begin{algorithmic}[1]
\Require Observation $o_t$, language instruction $l$, number of candidates $K$
\Require VLAT policy $\pi_\theta$, track-conditioned world model $p_\phi$, VLAC scorer $R_\psi$
\While{task not finished}
    \State $\{(a_i, \tau_i)\}_{i=1}^{K} \leftarrow \Call{Propose}{\pi_\theta, o_t, l, K}$
    \Comment{sample candidate action--track pairs}
    \For{$i = 1$ to $K$}
        \State $\hat{v}_i \leftarrow p_\phi(o_t, \tau_i)$
        \Comment{roll out future video with predicted tracks}
        \State $s_i \leftarrow R_\psi(\hat{v}_i, l)$
        \Comment{score predicted outcome}
    \EndFor
    \State $i^\star \leftarrow \arg\max_i s_i$
    \State Execute selected action chunk $a_{i^\star}$
    \State Observe new state $o_{t+H}$
    \State $t \leftarrow t + H$
\EndWhile
\end{algorithmic}
\end{algorithm}

\section{Pretraining Data and Unified Track Representation}

\paragraph{Unified Track Slot Layout.}
To enable joint pretraining on both robot and human manipulation datasets, we adopt a unified embodiment-agnostic track representation. We use \emph{slot} to denote a fixed index in a view-specific track-token layout, rather than a learned object slot. Each slot corresponds to a possible track type, such as a gripper/hand keypoint or a grid-sampled scene point. If a dataset does not provide the corresponding track type or camera stream, the slot is masked out. This fixed slot layout makes tracks a common interface across heterogeneous dataflows, where robot demonstrations with executable actions, human egocentric demonstrations without robot actions, and datasets with different camera configurations can all supervise the same track-prediction target. 

For DROID~\citep{khazatsky2025droidlargescaleinthewildrobot}, each agent-view camera contains 7 gripper-associated track points, while the wrist-view camera contains 7 gripper-associated track points together with 25 uniformly sampled background points.
For EgoDex~\citep{hoque2026egodexlearningdexterousmanipulation}, only a single egocentric-view camera is available, which contains 14 hand-associated track points (7 points per hand) together with 25 background points. Robot datasets activate only the subset of slots corresponding to the observed gripper or end-effector tracks, while the remaining slots are masked. This design allows observations from human demonstrations, single-arm robots, and future multi-manipulator embodiments with different camera-view setups to be represented within a shared token space. The two track groups provide complementary motion cues. Gripper- or hand-associated tracks capture EEF manipulator-centric motion, while uniformly sampled grid tracks provide scene and camera motion reference. Together, they help the model distinguish motion caused by the robot or hand from apparent motion induced by camera movement.

The main TrAct model is pretrained on DROID and EgoDex. In the appendix, we additionally study a larger pretraining mixture that includes BridgeData V2~\cite{walke2024bridgedatav2datasetrobot}. For BridgeData V2, each agent-view camera contains 7 gripper-associated track points, while the wrist-view camera contains 7 gripper-associated track points together with 25 background points. Since camera extrinsics are not available in BridgeData V2, we follow the same track extraction procedure used for DROID and obtain visual tracks directly from image observations via CoTracker.
The resulting tracks are represented using the same slot allocation and masking strategy as the rest of the pretraining mixtures, enabling seamless integration with both DROID and EgoDex data.

\paragraph{Track Tokenization and Masked Objective} Formally, each track token is represented as

\[
z_{t,v,i}
=
\phi(\tau_{t,v,i})
+
e_{\mathrm{view}}(v)
+
e_{\mathrm{slot}}(i)
+
e_{\mathrm{time}}(t),
\]

where $\tau_{t,v,i}$ denotes the track coordinate of slot $i$ from camera view $v$ at time $t$.
We introduce view-specific learnable embeddings to distinguish observations from different camera streams, while no embodiment-specific embeddings are used and the embeddings do not impose an embodiment-specific semantic meaning.

Missing manipulators, inactive track slots, unavailable camera streams, and unavailable action dimensions are handled through binary masks during attention and loss computation. Consequently, the model is encouraged to organize observations according to interaction dynamics rather than robot morphology or dataset-specific action formats, leading to embodiment-agnostic visual motion representations that transfer across embodiments and viewpoints.

For action prediction, we follow the same pretraining strategy while retaining embodiment-specific supervision.
Since executable robot actions and human hand motions do not share a common control parameterization, EgoDex is used only to supervise the track prediction head during pretraining.
Robot datasets additionally supervise both the track head and the action head using the corresponding robot action labels.
The overall training objective is

\[
\mathcal{L}
=
\mathcal{L}_{\text{track}}
+
\mathbb{1}_{\text{robot}}
\mathcal{L}_{\text{action}},
\]

where $\mathbb{1}_{\text{robot}}$ indicates whether robot action annotations are available.
Thus, tracks serve as a consistent supervision signal across all datasets, whereas actions are supervised only when their control format is valid and executable. This design allows TrAct to leverage large-scale human manipulation data to learn transferable visual motion representations while preserving robot-specific action supervision for downstream control.

\section{Inference Hyperparameters} 

We sample $K=20$ (simulation) and $K=16$ (real-world) candidates from TrAct at every inference step. These candidates are fed into the track-conditioned world model for action selection by VLAC. To maintain diverse candidates sampled from TrAct, we increase the noise scale and use temperature-scaled resampling (TSR). Our simulation experiments use a noise scale value of 2 and TSR k-value of 3. In our physical experiments, we use a noise scale of 2 and a TSR k-value of 4.

\begin{table}[!htb]
\centering
\small
\caption{Inference latency breakdown per planning step, measured with $K{=}16$ candidates, rolled out in parallel, on NVIDIA L40S GPUs. Each planning step yields a 16-step action chunk.}
\label{tab:latency}
\vspace{4pt}
\begin{tabular}{lcccc}
\toprule
\textbf{Component} & \textbf{VLAT} & \textbf{World Model} & \textbf{VLAC} & \textbf{Total} \\
\midrule
Latency (ms) & 52 & 1810 & 569 & 2431 \\
\bottomrule
\end{tabular}
\end{table}

\section{Additional Qualitative Video Comparison}
\begin{figure}[!htb]
    \centering
    \includegraphics[width=\columnwidth]{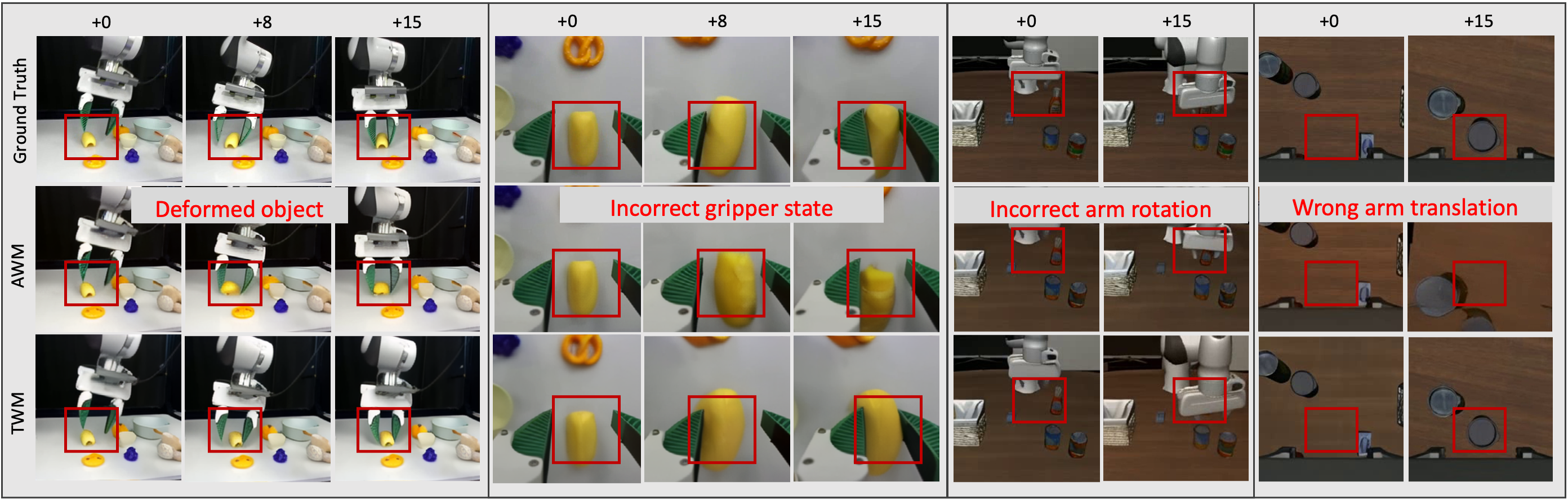}
    \caption{\textbf{Additional Qualitative video comparison.} TWM continues to produce more spatially accurate and motion-coherent frames than AWM across both agent-view and wrist-view cameras.}
    \label{fig:video_quality_comp_app}
\end{figure}

We provide additional qualitative comparisons between the track-conditioned world model (TWM) and the action-conditioned world model (AWM) in Figure~\ref{fig:video_quality_comp_app}. Consistent with the results in the main paper, TWM produces more spatially accurate and motion-coherent predictions across both agent-view and wrist-view cameras. By contrast, AWM more often exhibits spatial drift, distorted object shapes, and less coherent motion, especially in wrist view. These results further support that visual tracks provide a stronger conditioning signal than robot actions alone for future video prediction.

\section{Physical Experiment Setup}
\label{sec:physical_setup_supp}

\begin{figure*}[!htb]
    \centering
    \includegraphics[width=1\linewidth]{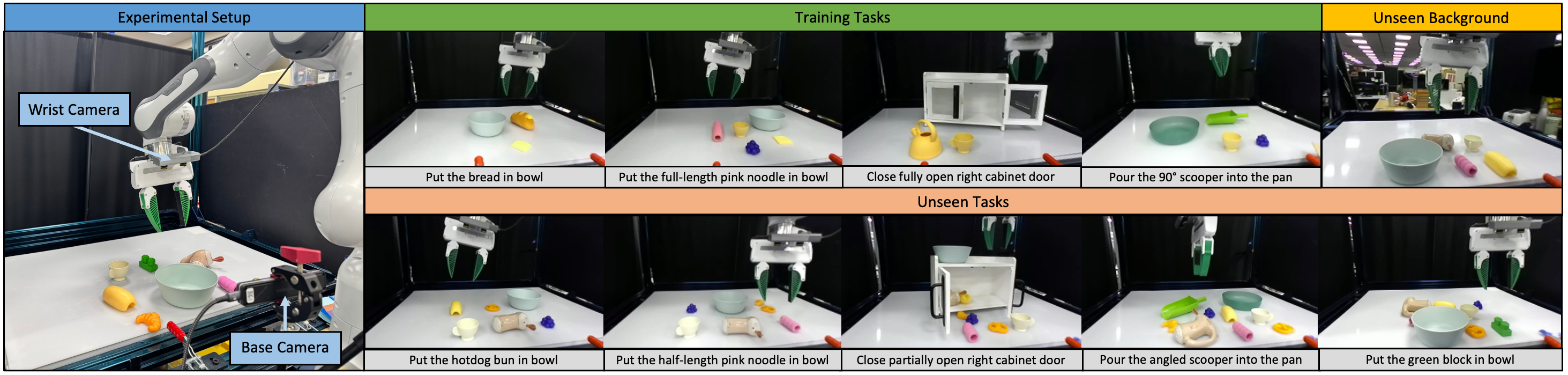}
    \caption{\textbf{Overview of the physical experiments.} We use a base and wrist-view camera and finetune the model on four training tasks. We then test five out-of-distribution tasks, as well as an unseen background for each task.}
    \label{fig:expsetup}
\end{figure*}

\begin{figure*}[!htb]
    \centering
    \includegraphics[width=1\linewidth]{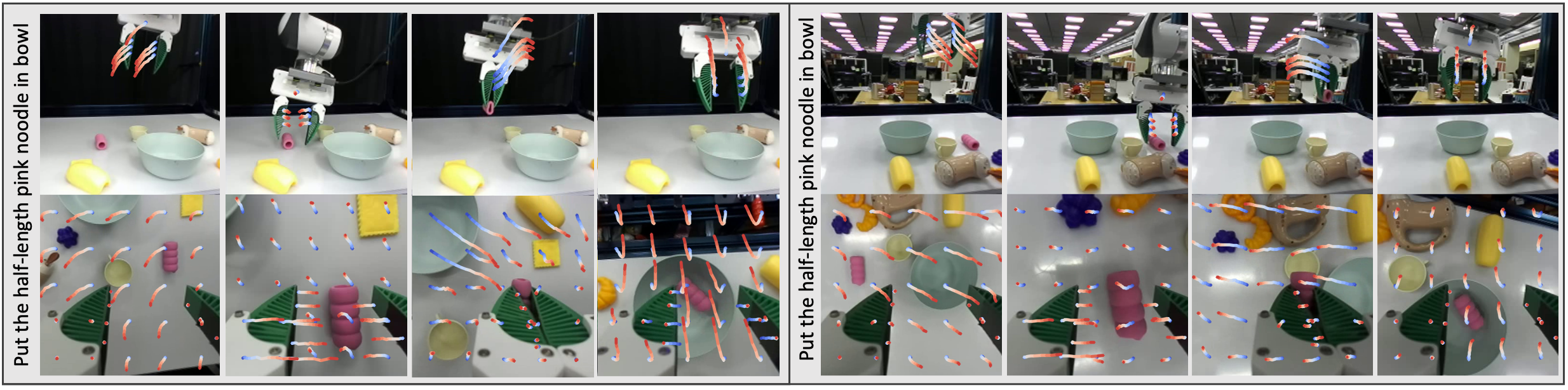}
    \includegraphics[width=1\linewidth]{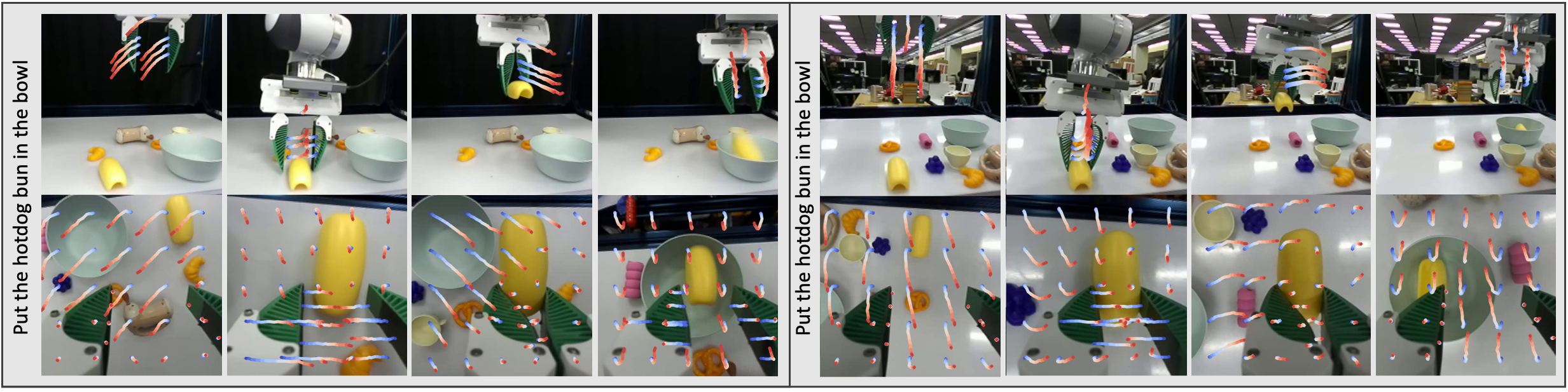}
    \includegraphics[width=1\linewidth]{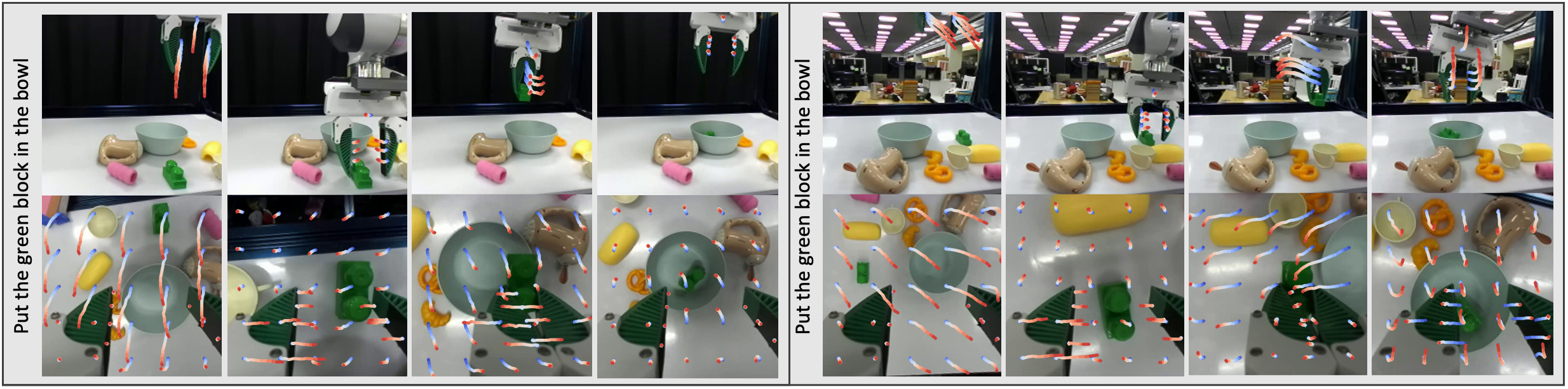}
     \includegraphics[width=1\linewidth]{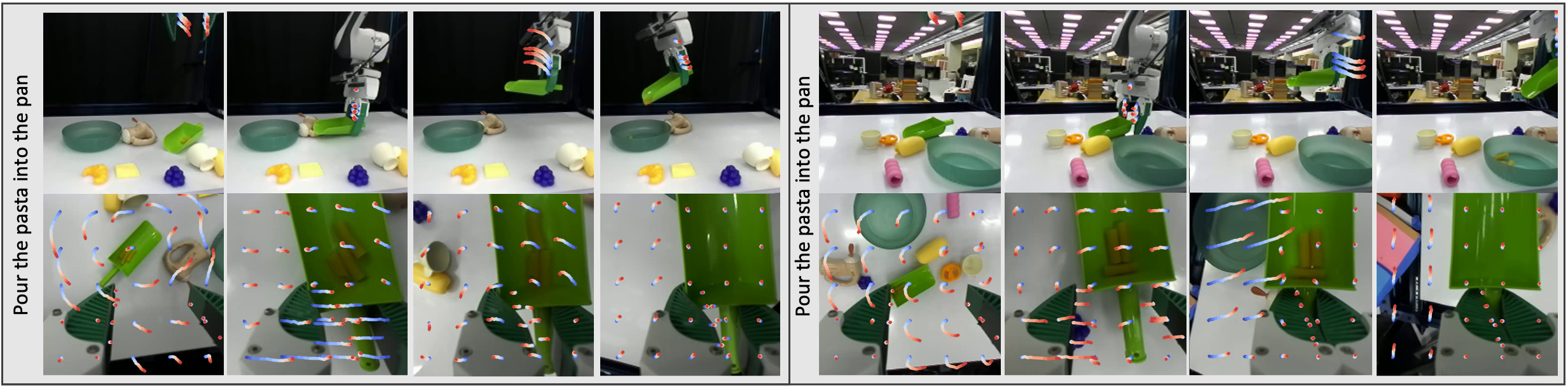}
     \includegraphics[width=1\linewidth]{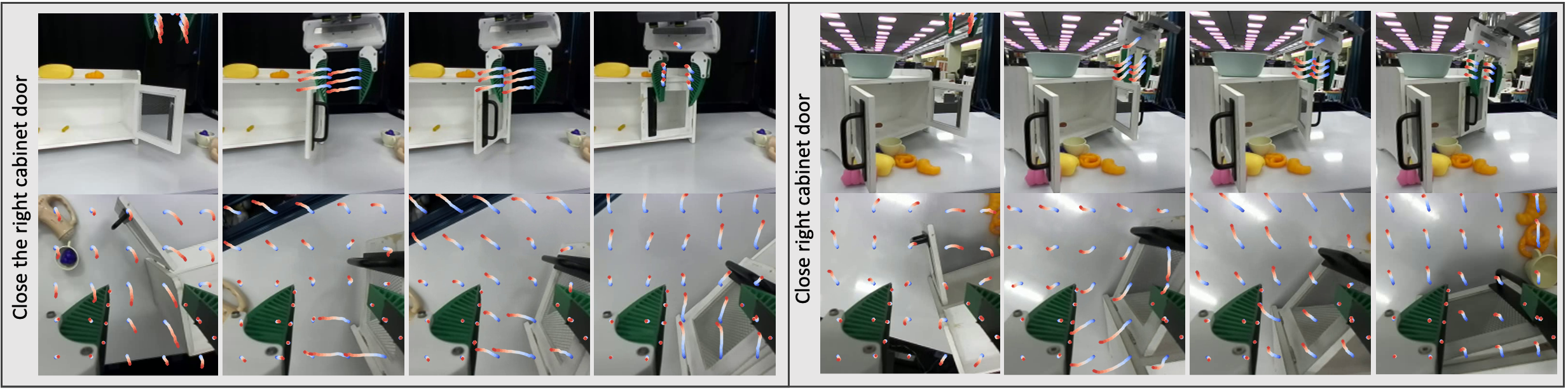}
    \caption{\textbf{Rollout examples} Rollout with four key frames showing 16-horizon track predictions under the training background (left) and unseen background (right). Task list from top row to bottom row: put the half-length pink noodle in the bowl (1), put the hotdog bun in the bowl (2), put the green brick in the bowl (3), pour the angled scooper's pasta into the pan (4), and close the partially open right cabinet door (5). }
    \label{fig:rollout_all}
\end{figure*}

We collected 400 demonstrations at 15 Hz for four tasks: put the pink noodle in the bowl, put the bread in the bowl, close the fully open right cabinet door, and pour the pasta into the pan with a fixed background. We evaluated five manipulation tasks that are unseen during training: put the half-length pink noodle in the bowl, put the hotdog bun in the bowl, put the green brick in the bowl, close the partially open right cabinet door, and pour the angled scooper's pasta into the pan. Each task is tested under two background conditions: the original training background (\texttt{reg.\ back}) and a visually distinct background (\texttt{diff back}), allowing us to measure robustness to visual domain shift on top of task-level generalization. The full setup is in Figure \ref{fig:expsetup}, and rollout examples with both background conditions are shown in Figure \ref{fig:rollout_all}.

\begin{table*}[p]
\centering
\footnotesize
\setlength{\tabcolsep}{5pt}
\renewcommand{\arraystretch}{1.0}
\caption{Results on the LIBERO-INTEGRAL benchmark. The benchmark includes robustness variations from LIBERO-PRO and LIBERO-Plus, and the last ten tasks evaluate cross-embodiment generalization by replacing the Franka Panda with a UR5 robot. Scores denote task success rates. The best result for each row is shown in bold.}
\label{tab:libero_integral_results_app}
\vspace{4pt}  
\begin{tabularx}{\textwidth}{>{\raggedright\arraybackslash}Xccccc}
\toprule
\textbf{Task} & \textbf{Var.} & $\boldsymbol{\pi_{0.5}}$ & \textbf{VLAT} & \textbf{VLAT+AWM} & \textbf{\algabbr{}} \\
\midrule

Pick the alphabet soup and place it in the basket
& Swap & 0.4 & 0.5 & \textbf{0.7} & \textbf{0.7} \\
\cmidrule(lr){1-6}

Put both the cream cheese box and the butter in the basket
& Swap & 0.4 & 0.5 & \textbf{0.6} & \textbf{0.6} \\
\cmidrule(lr){1-6}

Put both moka pots on the stove
& Object & 0.5 & 0.5 & 0.6 & \textbf{0.7} \\
\cmidrule(lr){1-6}

Put the yellow and white mug in the microwave and close it
& Object & 0.2 & 0.3 & 0.4 & \textbf{0.5} \\
\cmidrule(lr){1-6}

Pick the akita black bowl next to the ramekin and place it on the plate
& Task & 0.1 & 0.3 & \textbf{0.5} & \textbf{0.5} \\
\cmidrule(lr){1-6}

Put the plate on the stove
& Task & 0.2 & 0.4 & \textbf{0.5} & \textbf{0.5} \\
\cmidrule(lr){1-6}

Put the white mug on the left plate and put the yellow and white mug on the right plate
& Camera & 0.4 & 0.3 & 0.3 & \textbf{0.5} \\
\cmidrule(lr){1-6}

Put the wine bottle on the top of the drawer
& Camera & 0.5 & 0.6 & 0.6 & \textbf{0.7} \\
\cmidrule(lr){1-6}

Put both the cream cheese box and the butter in the basket
& RobotInit & 0.4 & 0.5 & 0.5 & \textbf{0.6} \\
\cmidrule(lr){1-6}

Pick up the book and place it in the back compartment of the caddy
& RobotInit & 0.5 & 0.6 & 0.6 & \textbf{0.7} \\
\cmidrule(lr){1-6}

Put both the alphabet soup and the cream cheese box in the basket
& UR5 & 0.1 & 0.2 & 0.2 & \textbf{0.3} \\
\cmidrule(lr){1-6}

Put both the alphabet soup and the tomato sauce in the basket
& UR5 & 0.3 & 0.6 & 0.6 & \textbf{0.7} \\
\cmidrule(lr){1-6}

Put the bowl on the stove
& UR5 & 0.2 & 0.3 & 0.3 & \textbf{0.5} \\
\cmidrule(lr){1-6}

Pick the orange juice and place it in the basket
& UR5 & 0.0 & 0.3 & 0.4 & \textbf{0.5} \\
\cmidrule(lr){1-6}

Pick the akita black bowl between the plate and the ramekin and place it on the plate
& UR5 & 0.1 & 0.5 & 0.5 & \textbf{0.6} \\
\cmidrule(lr){1-6}

Pick the akita black bowl next to the plate and place it on the plate
& UR5 & 0.1 & 0.3 & 0.3 & \textbf{0.5} \\
\cmidrule(lr){1-6}

Pick the ketchup and place it in the basket
& UR5 & 0.0 & 0.1 & \textbf{0.2} & 0.1 \\
\cmidrule(lr){1-6}

Pick the tomato sauce and place it in the basket
& UR5 & 0.0 & \textbf{0.2} & \textbf{0.2} & \textbf{0.2} \\
\cmidrule(lr){1-6}

Pick the akita black bowl on the cookies box and place it on the plate
& UR5 & 0.5 & \textbf{0.9} & \textbf{0.9} & 0.8 \\
\cmidrule(lr){1-6}

Put the bowl on the plate
& UR5 & 0.4 & \textbf{0.8} & \textbf{0.8} & \textbf{0.8} \\

\midrule
\textbf{Average}
& --
& 0.27 & 0.44 & 0.49 & \textbf{0.55} \\

\bottomrule
\end{tabularx}
\end{table*}

\section{Simulation Results}
\label{sec:appendix}

\subsection{Simulation Experiment Setup}
\label{sec:Sim_setup}

\begin{figure*}[!htb]
    \centering
    \includegraphics[width=1\linewidth]{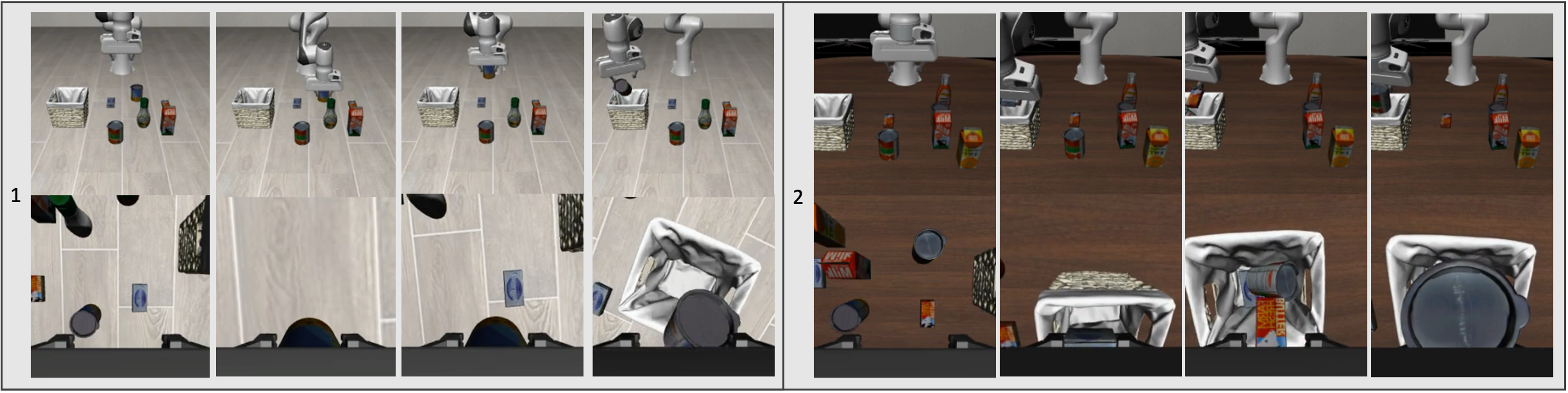}
    \includegraphics[width=1\linewidth]{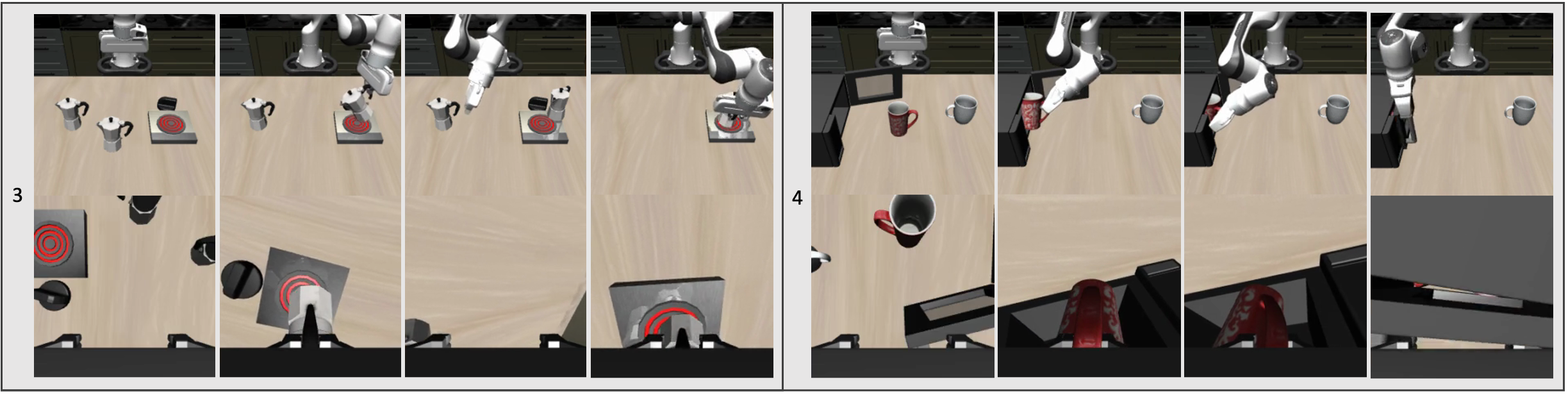}
    \includegraphics[width=1\linewidth]{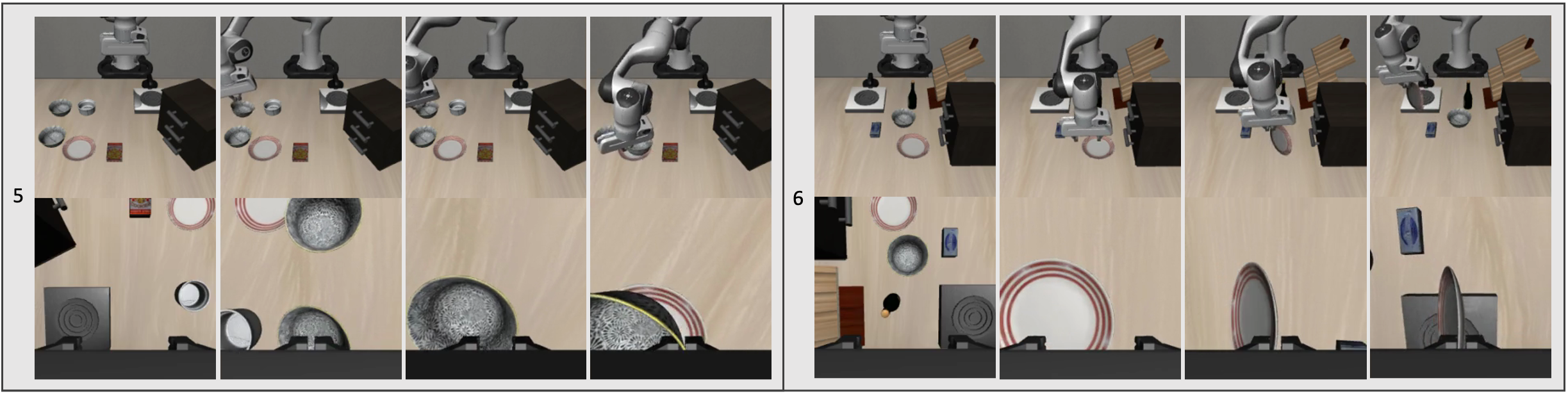}
     \includegraphics[width=1\linewidth]{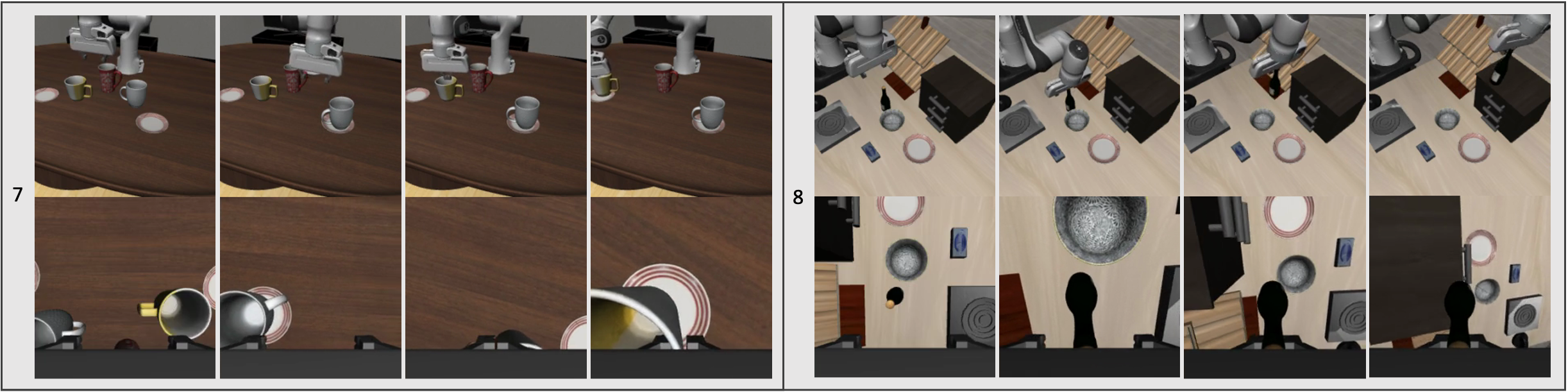}
     \includegraphics[width=1\linewidth]{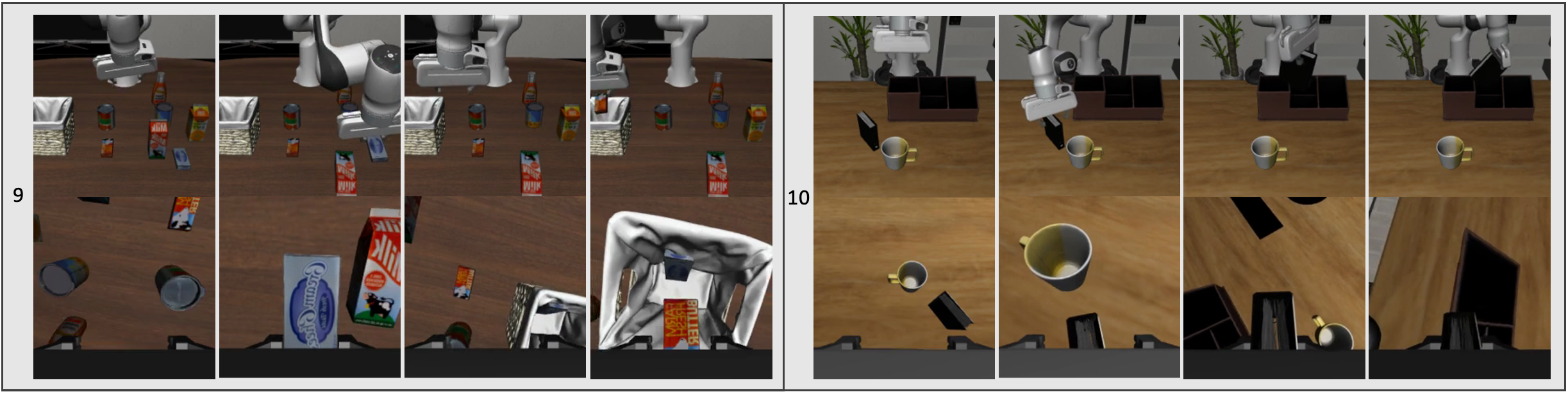}
    \caption{\textbf{Rollout examples} Rollout with four key frames. Task list from top row to bottom row, left to right column: pick the alphabet soup and place it in the basket (1), put both the cream cheese box and the butter in the basket (2), put both moka pots on the stove (3), put the yellow and white mug in the microwave and close it (4), pick the akita black bowl next to the ramekin and place it on the plate (5), put the plate on the stove (6), put the white mug on the left plate and put the yellow and white mug on the right plate (7), put the wine bottle on the top of the drawer (8), put both the cream cheese box and the butter in the basket (9), and pick up the book and place it in the back compartment of the caddy (10). }
    \label{fig:rollout_all_sim_1}
\end{figure*}

\begin{figure*}[!htb]
    \centering
    \includegraphics[width=1\linewidth]{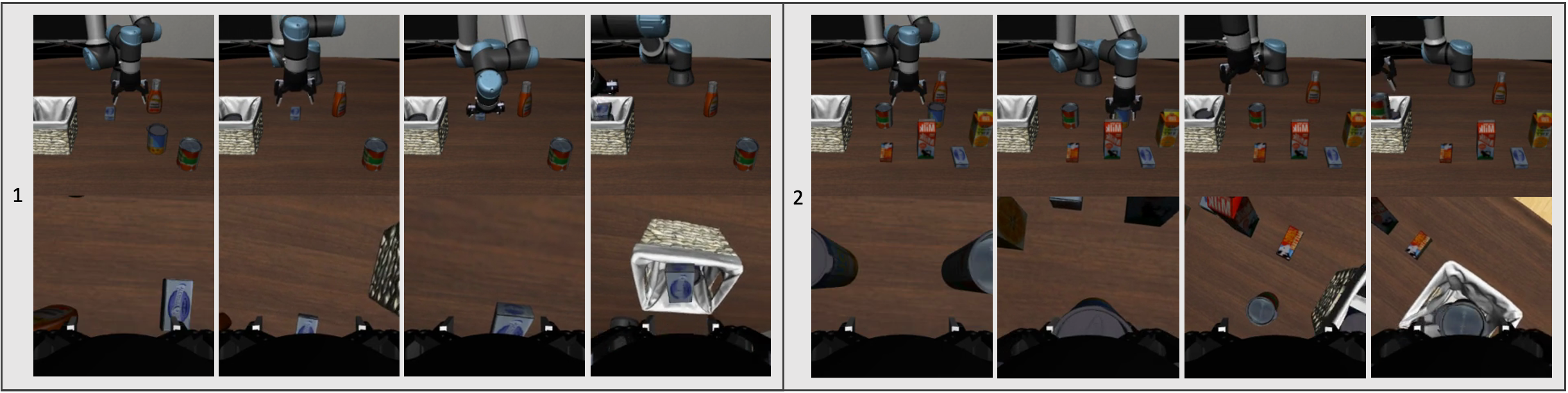}
    \includegraphics[width=1\linewidth]{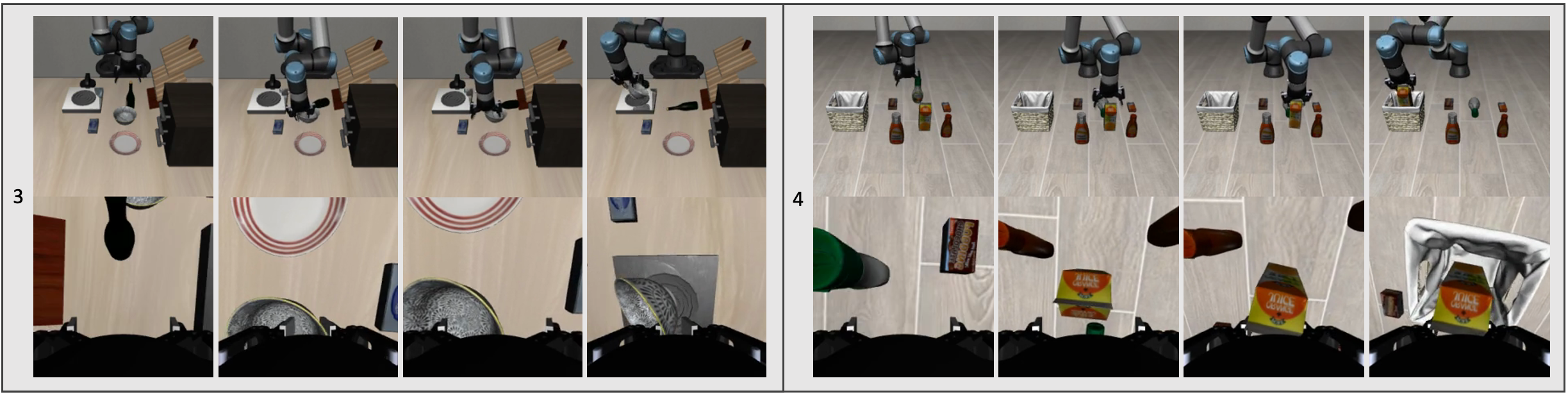}
    \includegraphics[width=1\linewidth]{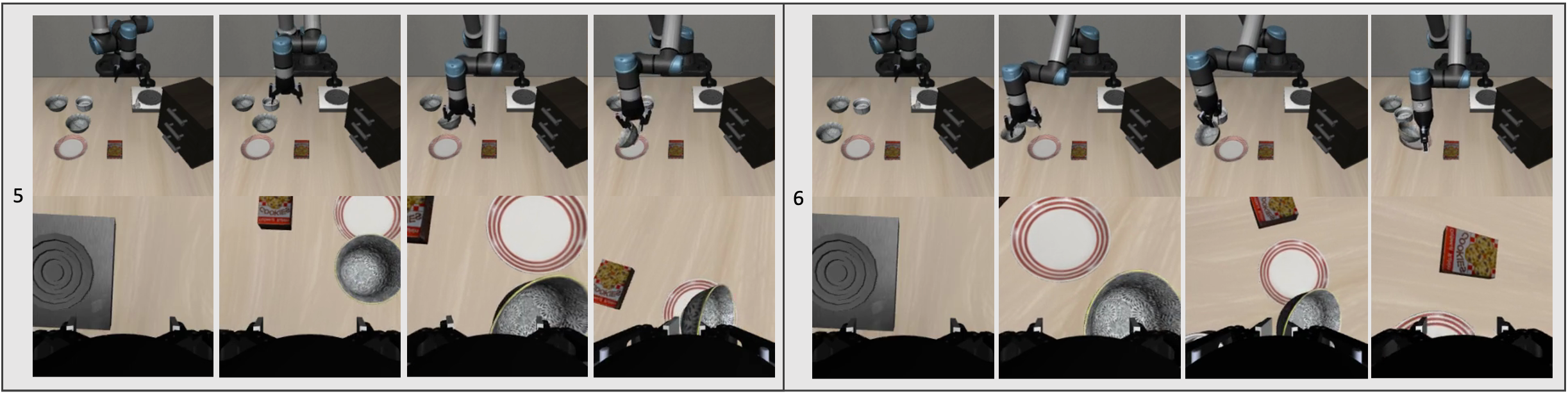}
     \includegraphics[width=1\linewidth]{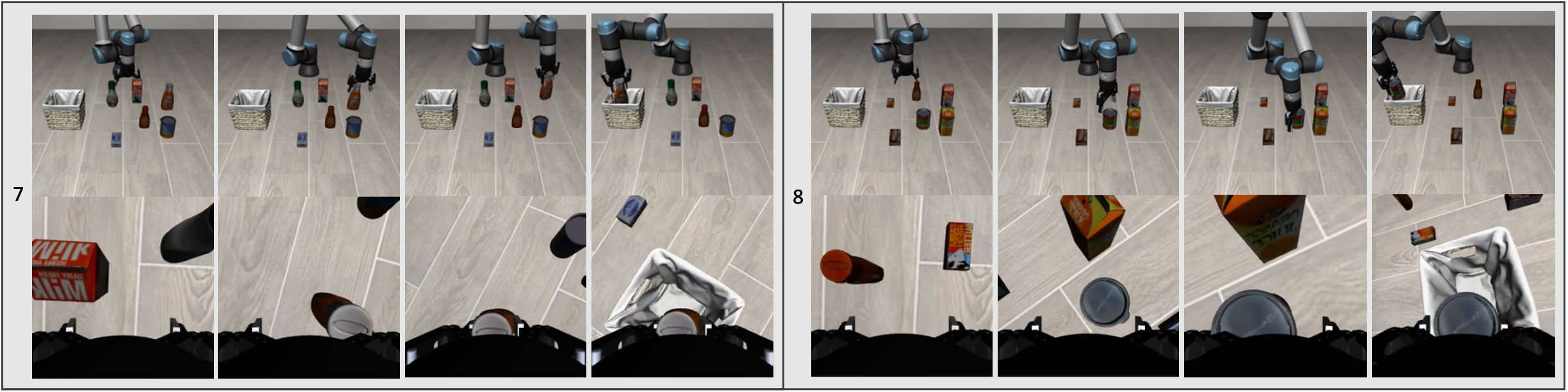}
     \includegraphics[width=1\linewidth]{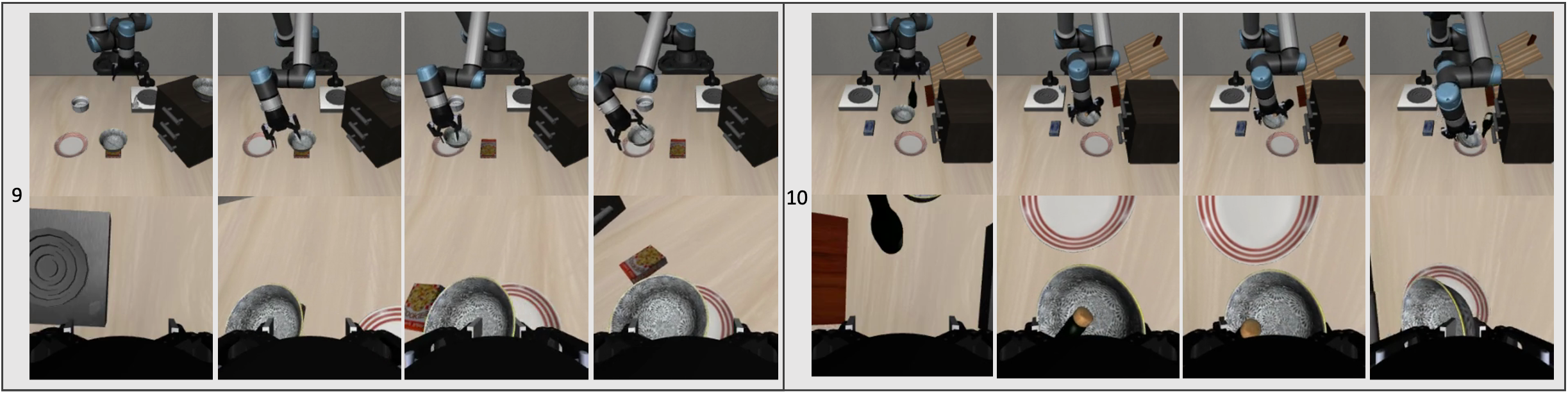}
    \caption{\textbf{UR5 Rollout examples} Rollout with four key frames. Task list from top row to bottom row, left to right column: put both the alphabet soup and the cream cheese box in the basket (1), put both the alphabet soup and the tomato sauce in the basket (2), put the bowl on the stove (3), pick the orange juice and place it in the basket (4), pick the akita black bowl between the plate and the ramekin and place it on the plate (5), pick the akita black bowl next to the plate and place it on the plate (6), pick the ketchup and place it in the basket (7), pick the tomato sauce and place it in the basket (8), pick the akita black bowl on the cookies box and place it on the plate (9), and put the bowl on the plate (10). }
    \label{fig:rollout_all_sim_2}
\end{figure*}

Since standard LIBERO benchmarks are nearly saturated, we introduce \textbf{LIBERO-INTEGRAL} to evaluate policy generalization under distribution and embodiment shifts. LIBERO-INTEGRAL includes 20 tasks: 10 robustness tasks and 10 cross-embodiment tasks. The robustness suite combines object, position, and task generalization from LIBERO-PRO~\cite{zhou2025libero} with the Different 3D Camera Viewpoints and Robot Initialization settings from LIBERO-Plus~\cite{fei2025libero}. We select two representative tasks per variation category and use the hardest available difficulty level for each LIBERO-Plus task. For cross-embodiment evaluation, we replace the default Franka Panda with a UR5 while keeping task specifications and scene configurations fixed. We construct UR5 variants of the 10 LIBERO-10 tasks and evaluate performance under this embodiment shift. We detail the full task selection in Figure \ref{fig:rollout_all_sim_1} and Figure \ref{fig:rollout_all_sim_2}.

\subsection{Per-Task LIBERO-INTEGRAL Results}
\label{sec:libero_integral_per_task}

To provide a more fine-grained view of the aggregate results in Table~\ref{tab:libero_integral_results}, Table~\ref{tab:libero_integral_results_app} reports the success rate of each individual task in LIBERO-INTEGRAL, including object swaps, object/task/camera variations, robot-initialization changes, and UR5 cross-embodiment transfer. \algabbr{} achieves the highest average success rate of 0.55, improving over $\pi_{0.5}$, VLAT, and VLAT+AWM by 0.28, 0.11, and 0.06 absolute points, respectively. Moreover, \algabbr{} achieves the best or tied-best performance on 18 out of 20 tasks, showing its consistent gains across diverse robustness and embodiment-shift settings.



\subsection{Statistical Robustness and Candidate Scaling}
\label{sec:seeds_and_k}

\begin{table}[!ht]
\centering
\small
\setlength{\tabcolsep}{3pt}
\begin{minipage}[t]{0.485\textwidth}
\centering
\caption{LIBERO-INTEGRAL success rates across three independent evaluation seeds. SE denotes the standard error over seeds and CI the corresponding 95\% confidence interval.}
\label{tab:appendix_seeds}
\vspace{4pt}
\begin{tabular}{lccc}
\toprule
\textbf{Method} & \textbf{Mean Success} & \textbf{SE} & \textbf{95\% CI} \\
\midrule
VLAT+AWM     & 0.490 & 0.006 & [0.47, 0.51] \\
\algabbr{}   & \textbf{0.547} & 0.003 & [0.53, 0.56] \\
\bottomrule
\end{tabular}
\end{minipage}\hfill
\begin{minipage}[t]{0.495\textwidth}
\centering
\caption{Effect of the number of sampled candidates $K$ on LIBERO-INTEGRAL average success rate, using VLAC selection over track-conditioned world model rollouts.}
\label{tab:appendix_k_sweep}
\vspace{4pt}
\begin{tabular}{lccc}
\toprule
\textbf{Selection Method} & $\boldsymbol{K=5}$ & $\boldsymbol{K=10}$ & $\boldsymbol{K=20}$ \\
\midrule
VLAC selection & 0.52 & 0.54 & \textbf{0.55} \\
\bottomrule
\end{tabular}
\end{minipage}
\end{table}

\paragraph{Multiple Seeds.}
The simulation results in the main paper are reported for a single evaluation seed. To verify that the gain of track conditioning over action conditioning is not an artifact of evaluation noise, we re-evaluate VLAT+AWM and \algabbr{} on the full LIBERO-INTEGRAL benchmark across three independent seeds, keeping all checkpoints, candidate counts, and selection hyperparameters fixed. Table~\ref{tab:appendix_seeds} reports the mean success rate, standard error, and 95\% confidence interval. The two intervals do not overlap, indicating that the improvement from track conditioning is consistent across runs rather than within-seed variance.

\paragraph{Number of Sampled Candidates.}
\algabbr{} selects among $K$ candidate action--track pairs at each planning step, so $K$ trades off compute against the quality of the best available candidate. We vary $K \in \{5, 10, 20\}$ on LIBERO-INTEGRAL while holding the policy, world model, and critic fixed. As shown in Table~\ref{tab:appendix_k_sweep}, success improves monotonically with $K$ but with diminishing returns, rising by 2 points from $K{=}5$ to $K{=}10$ and by a further 1 point from $K{=}10$ to $K{=}20$. Since world-model rollout cost grows with $K$ (Section~\ref{sec:conclusion}), $K{=}5$ already recovers most of the benefit of selection and is a reasonable operating point when inference latency matters; we use $K{=}20$ in the main simulation results to report the strongest achievable setting.


\section{Larger Pretraining Mixture Experiments}

\begin{table}[!ht]
\centering
\caption{
Additional results on the five most difficult UR5 cross-embodiment tasks from LIBERO-INTEGRAL. 
TrAct+ denotes TrAct pretrained with the larger DROID+BridgeData V2+EgoDex mixture. 
Scores denote task success rates. The best result in each row is shown in bold.
}
\label{tab:appendix_tract_plus}
\resizebox{\linewidth}{!}{
\begin{tabular}{lccc}
\toprule
\textbf{Task} & $\boldsymbol{\pi_{0.5}}$ & \textbf{TrAct} & \textbf{TrAct+} \\
\midrule
Put both the alphabet soup and the cream cheese box in the basket (UR5)
& 0.1 & \textbf{0.3} & \textbf{0.3} \\
Pick the ketchup and place it in the basket (UR5)
& 0.0 & 0.1 & \textbf{0.2} \\
Put the bowl on the stove (UR5)
& 0.2 & 0.5 & \textbf{0.6} \\
Pick the orange juice and place it in the basket (UR5)
& 0.0 & \textbf{0.5} & \textbf{0.5} \\
Pick the tomato sauce and place it in the basket (UR5)
& 0.0 & 0.2 & \textbf{0.3} \\
\midrule
\textbf{Average}
& 0.06 & 0.32 & \textbf{0.38} \\
\bottomrule
\end{tabular}
}
\end{table}

\paragraph{Larger Pretraining Mixture.}
We further evaluate whether TrAct benefits from scaling the real-world pretraining mixture beyond the DROID+EgoDex datasets used in the main paper.
Specifically, we pretrain a stronger variant, denoted as \textbf{TrAct+}, on a mixture of 76K DROID~\citep{khazatsky2025droidlargescaleinthewildrobot} trajectories, 60K BridgeData V2~\cite{walke2024bridgedatav2datasetrobot} trajectories, and the full EgoDex~\citep{hoque2026egodexlearningdexterousmanipulation} dataset, which contains approximately 300K processed egocentric dexterous manipulation clips.
Compared with the main model, which uses 76K DROID trajectories and a 150K subset of EgoDex,
this setting substantially increases both the diversity of robot embodiments and the coverage of human dexterous manipulation data.

We train TrAct+ for 60K steps with a global batch size of 128 on 8 H100 GPUs.\footnote{Unless otherwise specified, all batch sizes reported in this paper refer to the global batch size aggregated across all GPUs.}
For evaluation, we select the five most challenging UR5 cross-embodiment tasks from LIBERO-INTEGRAL, where TrAct obtains the lowest success rates, and compare $\pi_{0.5}$~\cite{intelligence2025pi05visionlanguageactionmodelopenworld}, TrAct, and TrAct+ under the same evaluation protocol.
As shown in Table~\ref{tab:appendix_tract_plus}, TrAct+ improves the average success rate from 0.32 to 0.38, demonstrating that \algabbr{} benefits from scaling both embodiment diversity and manipulation data coverage during pretraining.

\end{document}